\documentclass[letterpaper]{article}

\usepackage{amsmath}
\usepackage{amsfonts}
\usepackage{natbib,alifeconf}  
\usepackage{url,hyperref,cleveref}
\usepackage{booktabs}
\usepackage{caption}
\usepackage{subcaption}
\title{ALIFE2024 template}

\author{
    Jack Garbus$^{1}$,
    Thomas Willkens$^{1}$, 
    Alexander Lalejini$^2$, \and
    Jordan Pollack$^{1}$, 
    \mbox{}\\
    $^1$ Brandeis University, USA
    $^2$ Grand Valley State University, USA
    garbus@brandeis.edu
} 
\begin{document}

\title{Phylogeny-Informed Interaction Estimation Accelerates Co-Evolutionary Learning}

\maketitle
\begin{abstract}
  Co-evolution is a powerful problem-solving approach.
  However, fitness evaluation in co-evolutionary algorithms can be computationally expensive, as the quality of an individual in one population is defined by its interactions with many (or all) members of one or more other populations. 
  To accelerate co-evolutionary systems, we introduce phylogeny-informed interaction estimation, which uses runtime phylogenetic analysis to estimate interaction outcomes between individuals based on how their relatives performed against each other.
  We test our interaction estimation method with three distinct co-evolutionary systems: two systems focused on measuring problem-solving success and one focused on measuring evolutionary open-endedness.
  We find that phylogeny-informed estimation can substantially reduce the computation required to solve problems, particularly at the beginning of long-term evolutionary runs.  
  Additionally, we find that our estimation method initially jump-starts the evolution of neural complexity in our open-ended domain, but estimation-free systems eventually ``catch-up'' if given long enough.
  More broadly, continued refinements to these phylogeny-informed interaction estimation methods offers a promising path to reducing the computational cost of running co-evolutionary systems while maintaining their open-endedness.  
\end{abstract}
\hypertarget{introduction}{%
\section{Introduction}\label{introduction}}

Fitness prediction holds the potential to reduce the number of
evaluations required in a given generation of an evolutionary algorithm.
This benefit has motivated the development of fitness prediction methods, such as matrix factorization, neural estimation, fitness inheritance, and evolving fitness predictors \citep{schmidt_coevolution_2008,liskowski_surrogate_2016, bui_fitness_2005, pilato_fitness_2007, hiot_speeding-up_2010, liskowski_neural_2018}. 
However, few studies have investigated fitness prediction in co-evolutionary systems.
We introduce phylogeny-informed interaction estimation for co-evolutionary systems and investigate its efficacy for multi-population competitive co-evolution. 

A phylogeny (ancestry tree) describes the evolutionary history of an evolving population. 
Phylogenetic analyses that quantify evolutionary history are often applied post-hoc, providing useful insights into population-level evolutionary dynamics, such as diversification and long-term coexistence~\citep{dolson_quantifying_2018,lenski_evolutionary_2003, lalejini_evolutionary_2016}. 
Phylogenetic analyses have become increasingly easy to incorporate into evolutionary systems because of recent efforts to develop standardized formats for representing phylogenies~\citep{lalejini_data_2019} along with new software libraries for tracking phylogenies in a broad range of contexts~\citep{moreno_toward_2023}.
Indeed, recent work demonstrated the use of runtime phylogenetic analysis for fitness estimation in a single-population system~\citep{lalejini_phylogeny-informed_2023}. 


In single-population evolutionary search algorithms, estimating a candidate solution's fitness using the fitness of a nearby relative has been shown to reduce per-generation evaluation costs and improve problem-solving in some contexts~\citep{pilato_fitness_2007,hiot_speeding-up_2010,lalejini_phylogeny-informed_2023}.  
Co-evolutionary systems can also benefit from fitness estimation, as the fitness of an individual in one population is typically determined by evaluating it against many (or all) members of another population.
The cost of running such "all-versus-all" evaluations (``interactions'') increases quadratically as the co-evolving population sizes increase.
The computational costs of evaluation in co-evolutionary systems can reduce the scale at which we can apply them. 
Reducing the cost of evaluation would allow us to tackle bigger problems with co-evolutionary search algorithms and help us to scale up co-evolutionary artificial life systems to better support open-ended dynamics~\citep{taylor_open-ended_2016}.  

One way to reduce the computational cost of co-evolutionary fitness evaluation is to subsample the number of between-population interactions that are evaluated.
However, such subsampling may result in a significant loss of information needed to maintain diversity.
Furthermore, subsampling may prevent the application of state-of-the-art selection methods that require all-versus-all evaluations, such as the Discovery of Online Objectives (DISCO) algorithm~\citep{liskowski_online_2017}.
We investigate whether we can use estimation to support interaction subsampling for co-evolutionary algorithms that require all-versus-all evaluations, and we introduce a phylogeny-informed matchmaking method to improve expected estimate accuracy.



We tested our phylogeny-based estimation and matchmaking methods in the context of three co-evolutionary systems: sorting networks~\citep{hillis_co-evolving_1990}, the numbers game~\citep{watson_coevolutionary_2001}, and the collision game~\citep{willkens_evolving_2022}.
Overall, we found that phylogeny-informed methods can approximate the dynamics of all-versus-all evaluation while significantly reducing the number of required evaluations, particularly at the early stages of a run.
Consistent with other studies~\citep{lalejini_phylogeny-informed_2023}, the effectiveness of phylogeny-based estimation varies by domain, motivating future refinements to our methods.

\hypertarget{related-work}{%
\section{Related Work}\label{related-work}}



Subsampling is well-studied in the context of single-population evolutionary algorithms.
For example, random down-sampling, informed down-sampling, and cohort partitioning have been extensively tested in the context of the lexicase selection algorithm~\citep{spector_assessment_2012,hernandez_suite_2022, boldi_informed_2023}.
These methods can improve performance for a given compute budget by reallocating computational resources to increase the population size or
search time \citep{hernandez_random_2019}.
However, these subsampling methods cannot be directly applied to co-evolutionary algorithms without substantial modifications. 
Furthermore, methods like random subsampling have the potential to miss out on highly informative evaluations. 
For co-evolutionary algorithms, \citet{harris_competitive_2021} investigated using Elo as a metric to focus evaluations of individuals of similar skill, but found mixed results, with Elo often hurting the performance of competitive co-evolution in intransitive games.

A more general approach to subsample interactions for co-evolutionary systems involves evaluating a
fraction of interactions and estimating the rest. Estimation allows any
interaction subsampling method to work with any co-evolutionary algorithm
requiring the full interaction matrix \(G\), like DISCO~\citep{liskowski_online_2017}. Prior
methods for interaction estimation in evolutionary algorithms like
Surrogate Fitness via Factorization of Interaction Matrix (SFIMX)
extend non-negative matrix factorization to matrices with missing
entries \citep{liskowski_surrogate_2016}. Factor matrices \(W\) and
\(H\) can be calculated and then multiplied to reconstruct the
approximate interaction outcome matrix \(\hat{G} = WH\) with estimates
for missing entries based on the other entries in the matrix.

\citet{liskowski_neural_2018} propose Neural Estimation of Interaction Outcomes (NEIO), which predicts interaction outcomes using auto-encoders. Unlike SFIMX, this method can build non-linear models between interaction outcome, making it more
powerful than SFIMX at the cost of the significant systematic complexity
that comes with training and updating an auto-encoder alongside an
evolving population.

Others approaches co-evolve populations of fitness predictors alongside a focal solution population.
\citet{schmidt_coevolution_2008} co-evolve fitness predictors (maximizing prediction accuracy), solutions (maximizing predicted fitness), and fitness trainers (test cases that identify inconsistencies between predicted and true fitness).  
\citet{drahosova_adaptive_2019} also evolve a population of fitness predictors composed of subsets of training data for use in Cartesian Genetic
Programming. 
\citep{miller_designing_1999} show that fitness prediction reduces the time required for search with negligible impact on performance. 
These approaches, however, also dramatically increase the complexity of the evolutionary system as they evolve additional populations alongside the primary one, making them unwieldy to apply to systems that already include co-evolution.

A simpler approach to fitness estimation uses the outcomes of
related individuals to estimate members of the current population.
\citet{bui_fitness_2005} introduce a method of fitness inheritance for
sexually reproducing populations for use in NSGA2, where children
inherit the mean fitness of their parents and are only re-evaluated when an estimate falls outside some confidence interval. This approach is shown to be competitive to resampling methods with a lower computational cost and successfully reduces the time required to evolve hardware designs for Field Programmable Gate Arrays (FGPAs) by $25\%$ \citep{pilato_fitness_2007,hiot_speeding-up_2010}.

Recent work by \citet{lalejini_phylogeny-informed_2023} proposed phylogeny-informed
fitness estimation for genetic programming with lexicase selection,
where solutions inherit the score of their nearest ancestor or relative
for tests they are not evaluated on. Their method is shown to mitigate
the drawbacks of down-sampled lexicase methods and improve exploration
and diversity maintenance on some problems. This method, however, only
applies to problems with a static test set and does not introduce methods for choosing tests which improve estimator performance.

Little prior work has focused on estimating interactions outcomes in
pure co-evolutionary settings. Estimation techniques for evolutionary settings, such as NEIO, may not work in co-evolutionary settings, where both the problems and the solutions are changing. The only prior work we are aware of comes from \citet{arrojo_investigating_2018}, who
investigated co-evolutionary fitness estimation using Gaussian Processes, but these methods performed relatively poorly and exhibited
high degrees of variation in solution quality.

\hypertarget{method}{%
\section{Methods}\label{method}}

We define a phylogeny as a graph. 
Nodes in phylogeny are taxa, representing individuals that existed in the system at some point (or currently exist). 
Edges between nodes represent parent-child relationships. 
We compute the relatedness between two individuals in the same population from their distance in the phylogeny. 
Using this distance, we can then estimate the expected outcome of an interaction between two individuals given how the outcomes of interactions between their relatives.
Our approach can scale to $n$ co-evolving populations for up to $\frac{n(n-1)}{2}$ possible pair-wise interactions.

\hypertarget{defining-distances}{%
\subsection{Defining Distances}\label{defining-distances}}

To estimate the outcome of an interaction based on a related
interaction, we first measure the relatedness of the individuals in the
interaction.
For individuals, we can measure the pairwise distance
\(D_p(a_i,a_j)\) between individuals \(a_i\) and \(a_j\) from population
\(A\) as the shortest path between \(a_i\) and \(a_j\) on the
phylogenetic tree. For example, a parent and child would have a distance of one, while siblings would have a distance of two, as would a grandparent and grandchild. To compute a distance between two interactions
\(I_i,I_j\) across two populations \(A\) and \(B\), we need to
incorporate the pairwise distances between individuals and their
relatives from each population. We define an interaction between \(a_i\)
and \(b_j\) as \(I_{a_i,b_j}\), and the distance measure between two
interactions \(I_{a_i,b_j}\) and \(I_{a_m,b_n}\) as
\(D_I(I_{a_i,b_j}, I_{a_m,b_n})\). For our experiments, we use the following formula for computing interaction distance:

\begin{equation}
D_I(I_{a_i,b_j}, I_{a_m,b_n}) = D_p(a_i,a_m) + D_p(b_j,b_n)
\end{equation}
or, simply put, the sum of the distances between each pair of
relatives. An interaction between two parents would have a distance of two from an interaction between their children, whereas an interaction between two individuals would be one away from an interaction between one of those individuals and the other's parent.

\hypertarget{phylogeny-informed-interaction-estimation}{%
\subsection{Phylogeny-informed
Interaction Estimation}\label{phylogeny-informed-interaction-estimation}}

For simplicity, we describe our approach to phylogeny-informed estimation in the context of a two-population system.
Our method, however, can be easily scaled to N-population systems. 

To estimate an interaction between two individuals from different populations, we incorporate phylogenetic information from each individual.  
We compute our estimate as a weighted average of the \(k\)-nearest interaction outcomes. 
We define an interaction as a game played between two individuals, and an interaction outcome as the scores for each individual after the game has been played. 
We use the notation $I_x$ to denote both interactions and their outcomes, as context is sufficient to distinguish between the two.

First, we find the \(k\)-nearest interactions \(\mathcal{I}_i\) for a given
interaction \(I_i\) via a breadth-first search which iterates over pairs of nodes between the two trees. Next, we compute the estimated interaction outcome as the weighted average of
\(k\)-nearest interaction outcomes.
For a given set of \(k\)-nearest interactions \(\mathcal{I}_i\), we
define the total distance of \(\mathcal{I}_i\) as 

\begin{equation}
D_{\mathcal{I}_i} = \sum_{I_j \in \mathcal{I}_i} D_I(I_i, I_j)
\end{equation}

The weight of each evaluated outcome \(I_j \in \mathcal{I}\) on
the estimation of a different outcome \(I_i\) is then given by the complement of its distance
to the total distance of the interaction set: \begin{equation}
w_{i,j} = \frac{D_{\mathcal{I}_i} - D_I(I_i, I_j)}{D_{\mathcal{I}_i}}.
\end{equation}

Interaction outcomes are thus estimated as the weighted average of the $k$-nearest interactions: 
\begin{equation}
\mathbb{E}[I_{i}] = \sum_{I_{j} \in \mathcal{I}_i} w_{i,j}I_{j}
\end{equation}

where closer interactions have higher weights.

With this formulation, interactions that are distant relatives
contribute less to the estimate than interactions which are closely
related. We can then use the estimated outcomes and the evaluated
outcomes in any ordinary selection scheme. 

\hypertarget{phylogeny-informed-matchmaking}{\subsection{Phylogeny-informed Matchmaking}\label{phylogeny-informed-matchmaking}}

Phylogeny-informed Matchmaking chooses interactions to evaluate that are most expected to improve the accuracy of all outcome estimates. 
In this work, matchmaking strictly refers to the process of choosing which interactions to evaluate and which to estimate. 
Our use of the term matchmaking is not to be confused with the traditional notion of matchmaking where individuals of similar skill are paired together to create a more informative game.
We simply use the nomenclature "matchmaker" to convey the concept of pairing individuals against each other.

We assume that phylogenetic distance is negatively correlated with estimate accuracy,
so we want to choose the \(N\) interactions that minimize the average
distance between an interaction and its \(k\)-nearest evaluated
relatives.
There are \(\frac{(|A||B|)!}{(|A||B|-N)!}\) possible sets of
interactions to choose from.
Instead of computing the optimal set of interactions to evaluate which would maximize the expected accuracy of our estimates, we propose a simpler matchmaking scheme that guarantees low interaction distances for at least two of the $k$-interactions. We call this scheme \textit{parents-versus-all}, because we evaluate each individual in a population against all the parents of the opposing population. This method is particularly efficient in settings where a small number of parents have many children.

To measure the impact of parents-versus-all, we introduce a
random-cohort matchmaking scheme as a baseline. For this scheme, we divide each population into \(c\) random cohorts of a fixed
size, pair up each cohort with another cohort from a different
population, and run all-versus-all between paired cohorts. This randomly subsamples interactions while ensuring that each member of a
population is evaluated the same number of times. For each of the experiments shown, we ensure that the random cohort matchmaker samples
at least the same number of interactions as parents-versus-all, if not more.

\section{Experiments}
\subsubsection{Multi-dimensional Numbers Game}

The Numbers Game (NG) is a well-studied
evolutionary benchmark that has take many forms \citep{jong_ideal_2004, liskowski_online_2017, watson_coevolutionary_2001}.
For our purposes, we follow the implementations of
``CompareOnAll'' and ``CompareOnOne'' described in
\citet{liskowski_online_2017}. 
We evolve two populations of three-dimensional, real-valued vectors that we mutate by
adding uniform noise between -0.1 and 0.1 to two random dimensions during reproduction. A ``CompareOnAll'' interaction outcome
for vector \(a\) when playing against \(b\) is as follows:
\begin{equation}
g_\text{all}(a,b) = \begin{cases} 
1 & \text{if } \forall_{i=1,2,3} \text{ } a_i \geq b_i\\
0 & \text{otherwise}
\end{cases}
\end{equation}

whereas a ``CompareOnOne'' Interaction outcome only scores \(a\) on \(b\)'s largest dimension, \(j = \text{argmax }(b)\):

\begin{equation}
g_\text{one}(a,b) = \begin{cases} 
1 & \text{if } a_j \geq b_j\\
0 & \text{otherwise}
\end{cases}
\end{equation}

For this domain, we use fitness-proportional selection, which computes fitness as the average outcome over
all evaluated and estimated interactions.
We chose the Numbers Game because it is fast to evaluate, simple to understand, and has a smooth fitness landscape. 
As such, an the relatives of interacting individuals should serve as good estimates of the outcome, allowing us to test our estimation and matchmaking methods under ideal conditions. 


\subsubsection{Sorting Networks} 
A sorting network is a sequence of comparison operations that sort a sequence of numbers. 
Seminal work in co-evolutionary research evolved sorting networks in competition with sets of numbers called parasites \citep{hillis_co-evolving_1990}.
Where \citet{hillis_co-evolving_1990} implemented a toroidal grid to mediate interactions between networks and parasites, our work evaluates (or estimates) all networks against all parasites. 
In addition, our networks reproduce asexually, as we leave estimation between sexually reproducing populations for future work.

We represent 16-input sorting networks as a variable-length list of pairs of numbers between 1 through 16.
Each pair defines a compare-exchange operation, called a swap, which specifies two inputs to compare and exchange if out of order. 
We mutate sorting networks by randomly adding swaps, removing swaps, moving swaps, or randomizing the positions a swap compares with 25\% probability for each operation. 
Each parasite is represented as a length-16 vector of integers, and is mutated by randomly switching two elements in the vector. 

To ``run'' a sorting network on a parasite, we apply each compare-exchange operation specified by the sorting network in order. 
If a network perfectly sorts all parasites, we add a bonus term to the network's fitness interpolated between 0 and 1 depending on the size of the network. 
Networks with 60 swaps get an extra unit of fitness, networks with 120 swaps get 0, and networks between this range get a value between 0 and 1 inversely proportional to the number of swaps. 
This adds pressure for networks to shrink in size once they can sort all inputs. 
For this domain, we test our estimation method using lexicase selection \citep{spector_assessment_2012},
an algorithm that requires the full interaction matrix. 




\subsubsection{The Collision Game}  
The Collision Game (CG) \citep{willkens_evolving_2022} is a two-player game where agents, controlled by dynamically sized neural networks, move left or right on a one-dimensional number line. The players are rewarded or punished for colliding
with their opponent depending on which population their opponent comes
from. For two-population settings where a ``host'' plays against either
a ``mutalist'' or ``parasite'' population, the optimal strategy for the host is to always
collide or always retreat from the opponent. Three-population configurations play hosts against both mutualist AND parasite populations. In these settings, the host cannot
be certain whether it is playing against a mutualist or parasite, an
arms race begins---the host must get better at differentiating between mutualists and parasites, and both the mutualist and the parasite must get better at convincing the host they are each a
mutualist. Unlike the other domains, there is no measure of ``objective'' fitness in the Collision Game. Progress is instead measured by increases in neural complexity as strategies become more sophisticated. We refer the reader to \citep{willkens_evolving_2022} for helpful visuals and additional information.

Prior research on this domain demonstrated unbounded neural
complexity growth under DISCO selection \citep{liskowski_online_2017}, so we use this domain to investigate whether estimation impedes the generation of complexity
This is of particular interest for evolutionary approaches to open-endedness, as estimation can potentially accelerate long-running open-ended experiments if the overall rate of complexification is not significantly reduced. 

We use the Generalized Acquisition of Recurrent Links algorithm
\citep{angeline_evolutionary_1994} to evolve the architecture and
weights of neural networks. We mutate networks by adding/deleting nodes and edges and by applying a
small amount of noise to all edge weights. We call a network \emph{minimized} when we remove all nodes and connections that do not contribute to the output, and we refer to the \emph{complexity} of a neural network as the number of connections in its minimized version.
\begin{figure*}
     \centering
     \begin{subfigure}[b]{\columnwidth}
         \centering
         \includegraphics[width=\linewidth]{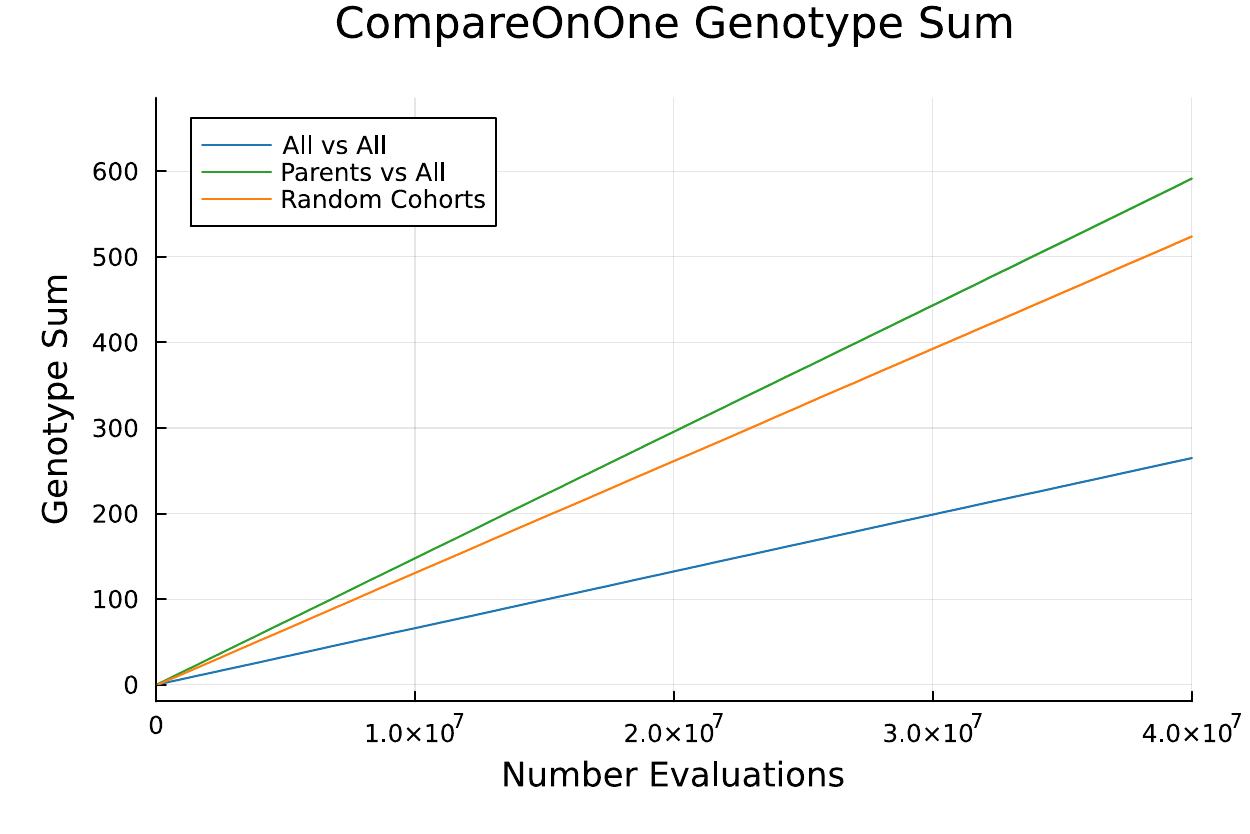}
         \label{fig:ng-coo-genotypesum}
     \end{subfigure}
     \hfill
     \begin{subfigure}[b]{\columnwidth}
         \centering
         \includegraphics[width=\linewidth]{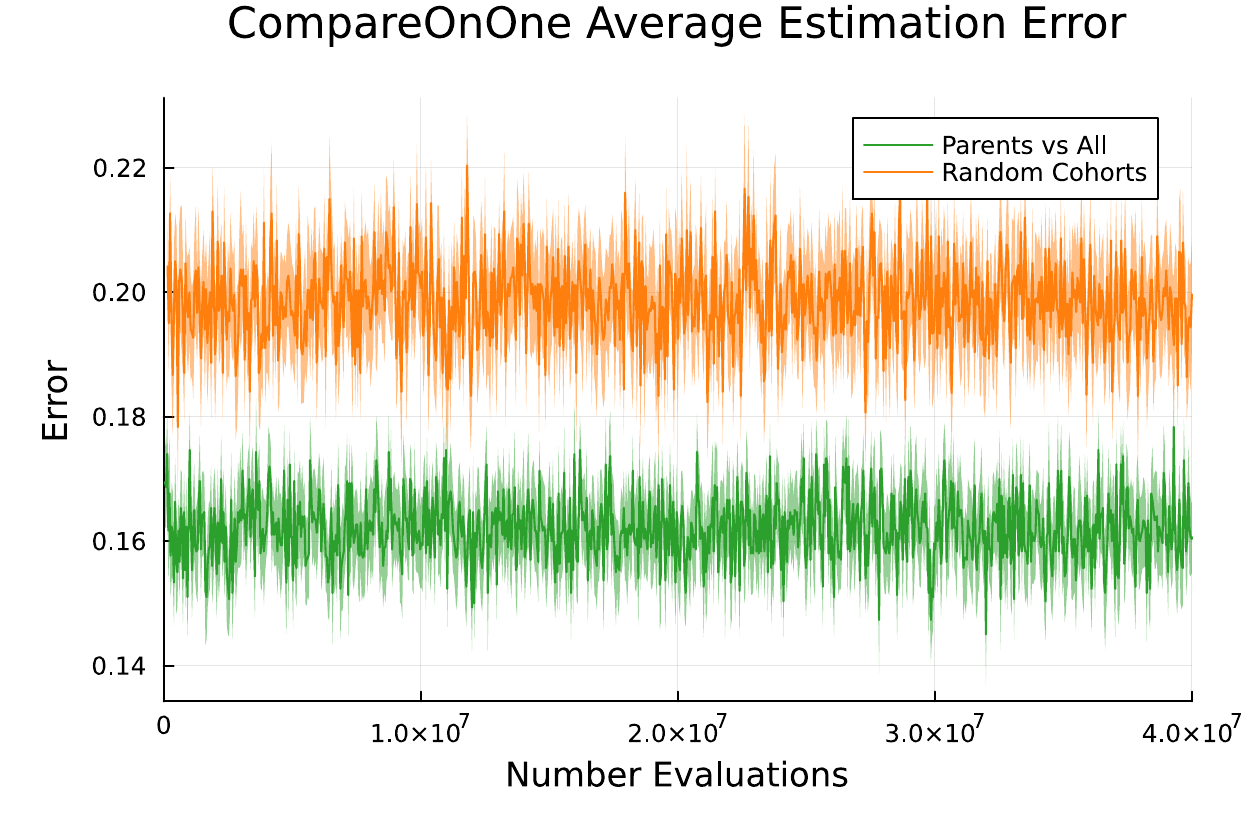}
         \label{fig:ng-coo-error}
     \end{subfigure}
     \caption{Results for the CompareOnOne setting of the Numbers Game across thirty trials, 95\% confidence intervals are shown, but very small. Left: Mean genotype sum across both populations for all three matchmaking methods. Right: Average estimation error between our two match making methods. The possible error of interaction estimation is bounded between 0 and 1, where 0 is perfect accuracy. Parents-versus-all results in consistently less error on this domain compared to random cohorts ($p\ll0.001$, Wilcoxon test; Glass's $\Delta=-1.15$).}
    \label{fig:ng-coo}
\end{figure*}

\hypertarget{experimental-setup}{
\subsection{Experimental Setup}\label{experimental-setup}}

For each domain, we compared three evaluation treatments: parents-versus-all with estimation, random cohort partitioning with estimation, and all-versus-all. Relevant hyperparameters can be found in Table \ref{tab:hparam}.
For all domains, we truncate the population to the specified number of parents before selecting individuals to reproduce.

We terminate search after finding $k$ interactions. If search exhausts all interactions within ten edges, we perform estimation using the weighted average of the interactions found so far. Search and estimation logic run quickly in constant time, independent of phenotype, making estimation particularly effective as phenotypes grow in size.

We additionally evaluate 200 random child-versus-child interactions, which we compare to their estimates to measure estimation error across matchmaking methods. These interactions are not used when computing estimates. We run thirty trials per treatment and run at least as many random cohort match-ups per generation as we do for parents-versus-all. All experiments use the $k=2$ nearest-interactions for estimation, implying that for parents-versus-all, child versus child matches are estimated using both parent-versus-child outcomes, which have an interaction distance of one. 
The random cohorts regime has no guarantees on how far a related interaction may be.
Preliminary experiments displayed no observable difference with $k>2$, as close parent interactions out-weighed distant ancestral interactions, so we use $k=2$ across both methods for efficiency.

We define an evaluation as the computation of an interaction outcome between two individuals.
All figures shown plot a metric against the number of evaluations, as we are trying to maximize evolutionary progress in the smallest amount of evaluations.  For example, running an all-versus-all matchmaker on two populations of size 100, will perform 10,000 evaluations per generation while a random cohort matchmaker with cohort size 50 only runs 5,000.

Due to competition for compute, we encountered issues accurately measuring wall-clock runtime. Trials often paused to provide other researchers with resources, and restoring from checkpoints in Julia requires significant re-compilation, which affects runtime. As configurations which take longer to run experience more interruptions, we do not explicitly detail end-to-end runtime statistics, but instead report approximate differences in time spent on evaluation and estimation during a generation.

All figures show bootstrapped 95\% confidence intervals around mean values. 
We use Kruskal-Wallis tests to assess statistical significance between the three treatments, and we use post-hoc Wilcoxon rank-sum tests for pairwise comparisons between treatments. 
To correct for multiple comparisons, we use a Bonferroni correction where appropriate. We include Glass's $\Delta$ to measure effect size. Unless otherwise specified, all statistical tests were performed on measurements made at the end of each run.

\hypertarget{results}{%
\section{Results}\label{results}}

\hypertarget{numbers-game}{%
\subsection{Numbers Game}\label{numbers-game}}

\begin{figure*}
\centering
\begin{subfigure}[b]{\columnwidth}
\centering
\includegraphics[width=\columnwidth]{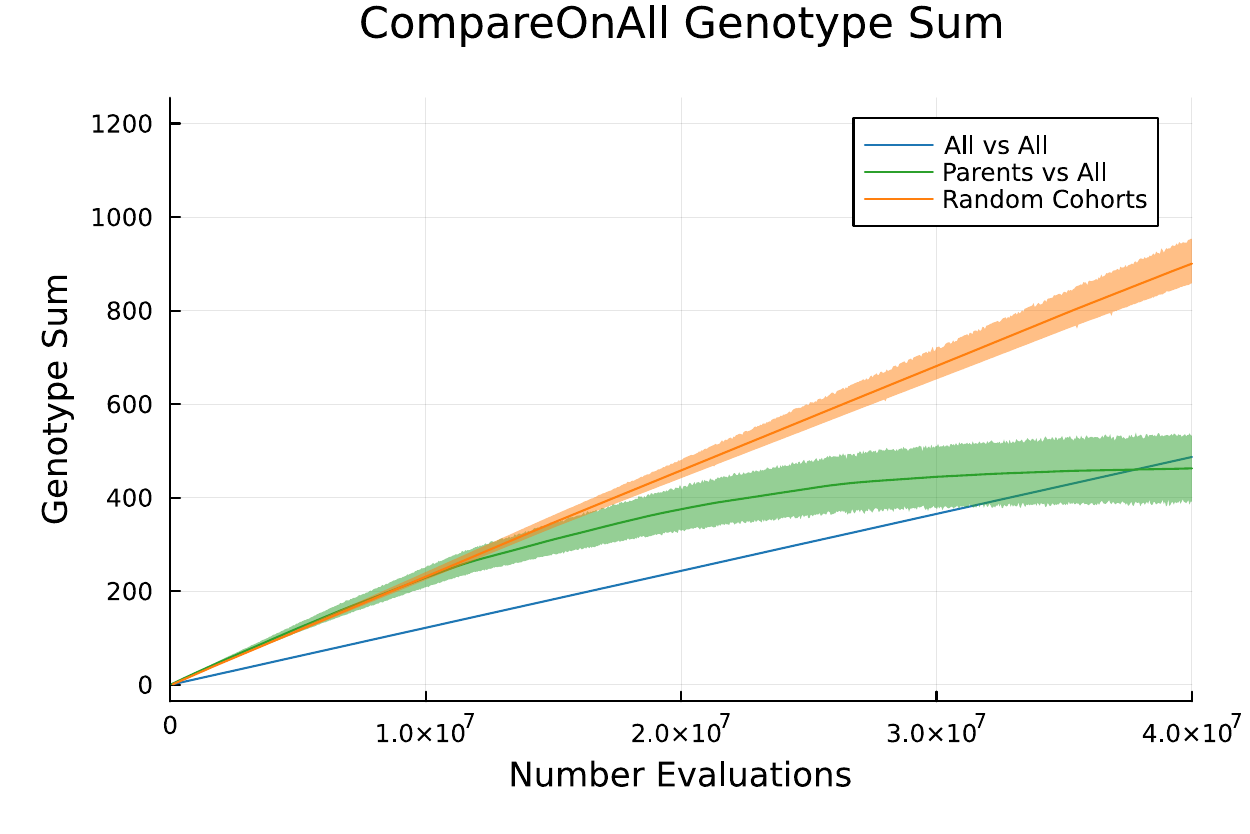}
\end{subfigure}
\begin{subfigure}[b]{\columnwidth}
\centering
\includegraphics[width=\columnwidth]{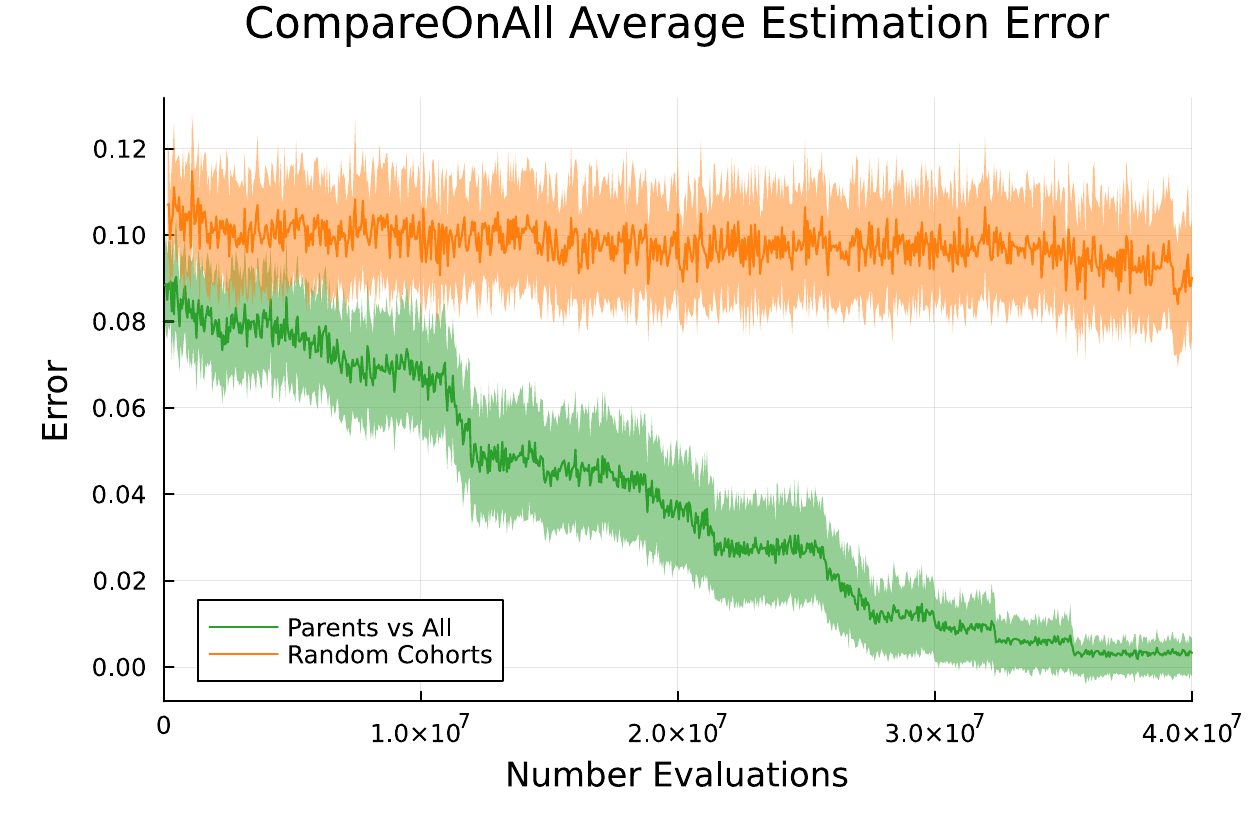}
\end{subfigure}
\caption{Results for the CompareOnAll setting of the Numbers Game across thirty trials, 95\% confidence intervals shown. \textbf{Left}:
The average sum of all dimensions over all genotypes for each
population. While all methods appear to perform well initially, the progress of parents-versus-all plateaus, unlike random cohorts. \textbf{Right}:
Estimation error in the CompareOnAll setting. Parents-versus-all matchmaking
results in lower average error than random cohorts ($p < 0.05$ at 2e6 evaluations, Wilcoxon test; Glass's $\Delta=-0.36$), even when
performing worse.}
\label{fig:ng-coa}
\end{figure*}

Figure~\ref{fig:ng-coo} shows the average genotype sum and estimation error for CompareOnOne, and Figure~\ref{fig:ng-coa} shows the average sum and error for CompareOnAll. 
We see stark differences in performance between the two NG domains. 
In Figure \ref{fig:ng-coo}, all three matchmaking methods continuously evolved increasingly high sums in the CompareOnOne configuration. 
For an equivalent number of evaluations, parents-versus-all outperformed both all-versus-all and random cohorts on CompareOnOne (non-overlapping 95\% confidence intervals). For both CompareOnOne and CompareOnAll, estimation does not significantly reduce evaluation time, as evaluation on these problems is always inexpensive. This domain, however, reveals an important property intrinsic to the parents-vs-all matchmaking method.  

In the CompareOnAll configuration, the random cohort matchmaking method accelerated co-evolution the most ($p\ll 0.0001$, Wilcoxon test; Glass's $\Delta=72.2$), and parents-versus-all appears to fizzle out (Figure~\ref{fig:ng-coa}). 
Analysis indicates that this is a domain-specific issue---the gradual decline in growth appears to be due to populations ``disconnecting'' one trial at a time, as previously seen in \citep{watson_coevolutionary_2001}. In this setting, it is possible for a population's members to evolve so much higher than their opponents across all dimensions that the greater population dominates the lesser population for every single interaction, making selection uniformly random. 
When the population vectors become too far apart for random drift to reconnect them, growth of all dimensions stagnate for the rest of the trial, indicating that parents-versus-all introduces some bias in the CompareOnAll domain. This phenomenon does not occur with random cohorts or all-versus-all, indicating that random-cohorts is a relatively bias-free estimation technique, even though it has higher estimation error (Figure \ref{fig:ng-coa}). 
We do not observe this disconnection phenomenon on any other domains in this study.

We hypothesized that the absence of child-versus-child matches led to the disconnect found in the CompareOnAll problem. 
We tested two potential mitigations to this phenomenon: (1) Each generation, choosing the parents-versus-all matchmaking algorithm with 95\% chance and all-versus-all with 5\% chance; and (2) for each child, instead of running $P$ match-ups against all $P$ parents, run $P-c$ match-ups against randomly selected parents and $c$ match-ups against randomly selected children, such that all parents play the same number of games and all children play the same number of games. 
Replacing random parent versus child matches with child versus child matches still results in population disconnects, but the disconnects take longer to occur proportional to increases in $c$. 
While outside the scope of this study, future work will investigate methods for mitigating population disconnects triggered by interaction estimations. 

\begin{figure}
\centering
\begin{subfigure}[b]{\columnwidth}
\includegraphics[width=\columnwidth]{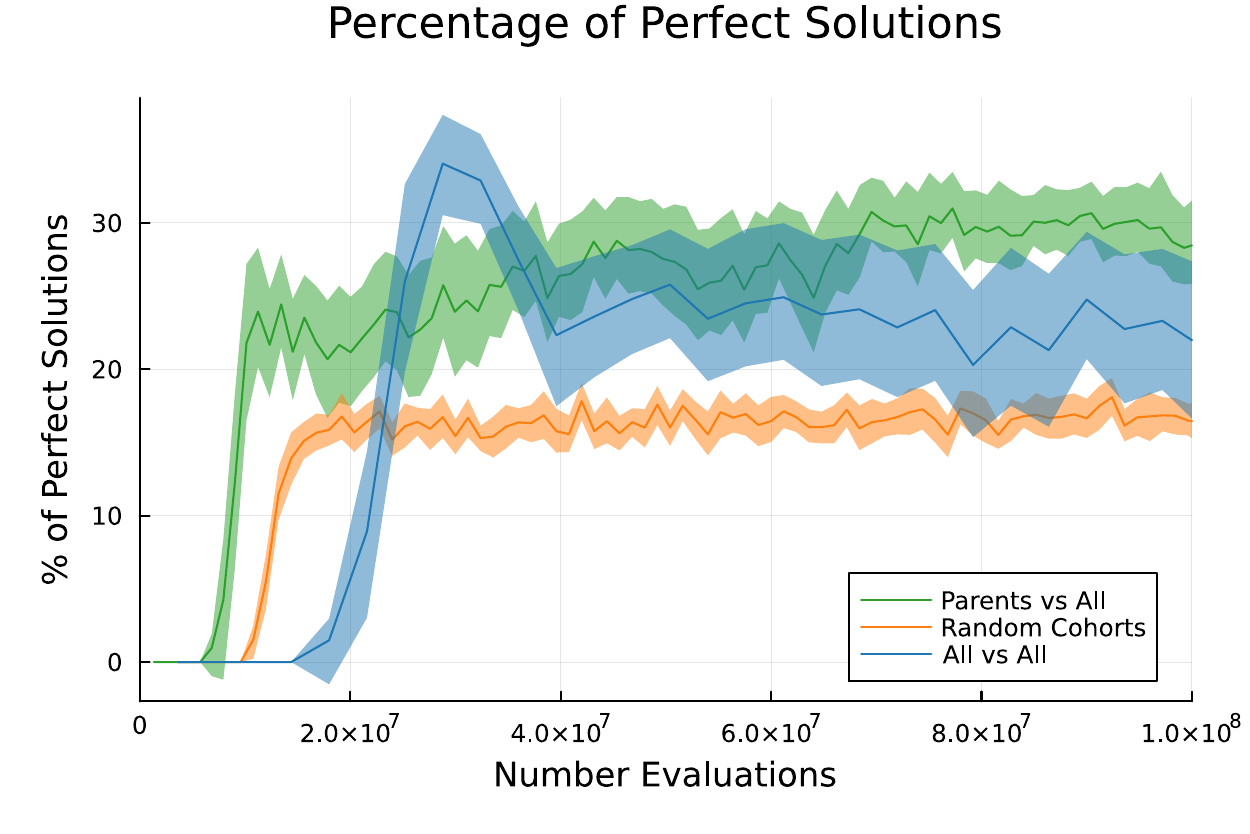}
\end{subfigure}
\begin{subfigure}[b]{\columnwidth}
\includegraphics[width=\columnwidth]{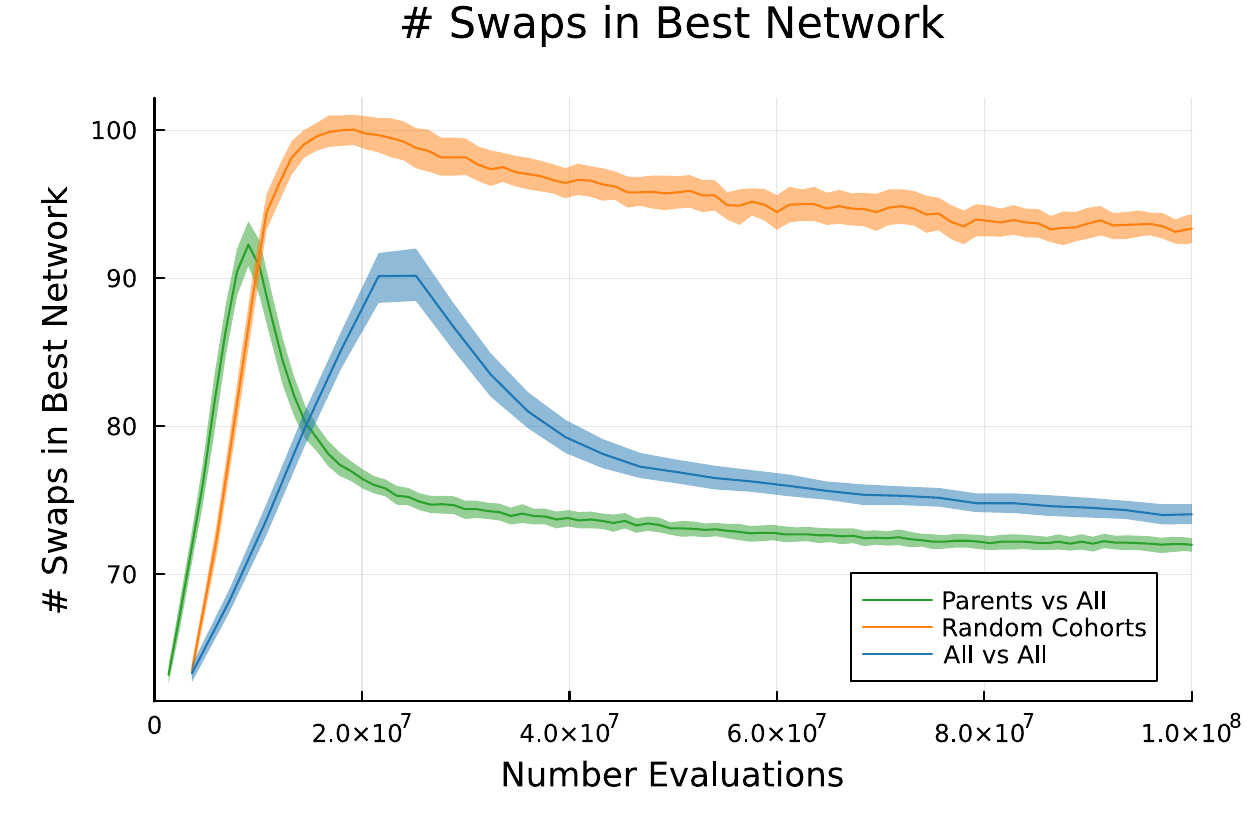}
\end{subfigure}
\begin{subfigure}[b]{\columnwidth}
\includegraphics[width=\columnwidth]{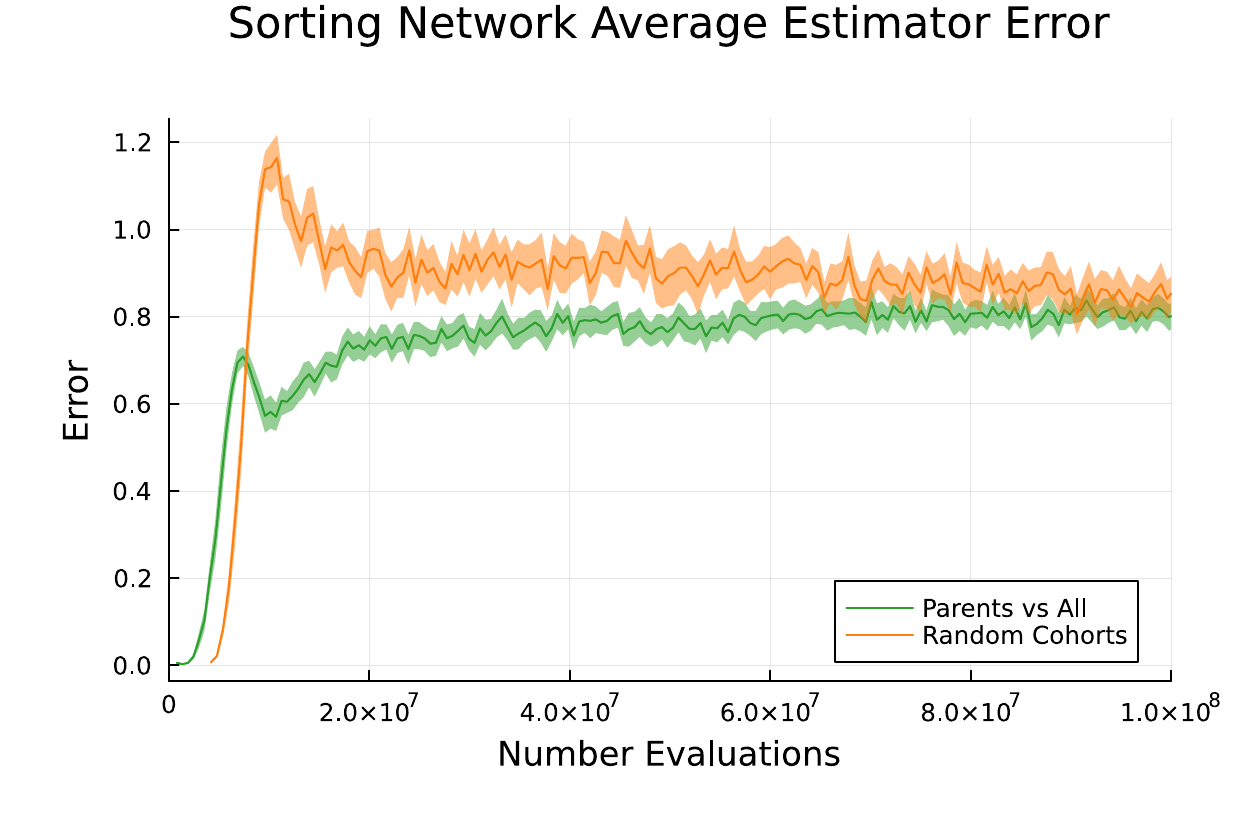}
\end{subfigure}
\caption{Results for Sorting Networks across thirty trials, 95\% confidence intervals are shown. \textbf{Top}: Average percentage of perfect sorting networks. \textbf{Middle}: Average number of swaps of the best sorting network. \textbf{Bottom}: Error of our estimators on 16-Input Sorting Network.  The minimum possible error is 0, and the maximum possible error is 16. The parents-versus-all method has significantly lower error than random cohorts during initial stages of evolution ($p\ll0.0001$ at 1e7 evaluations, Wilcoxon test; Glass's $\Delta=-2.42$).}\label{fig:sn}
\end{figure}

\hypertarget{sorting-networks}{%
\subsection{Sorting Networks}\label{sorting-networks}}
Parents-versus-all significantly outperforms random cohorts at minimizing perfect networks (Figure \ref{fig:sn}; $p\ll0.0001$, Wilcoxon test; Glass's $\Delta=-7.6$).
While both methods quickly solve the task faster than all-versus all by saturating the networks with swaps, parents-versus-all shrinks network sizes in a fashion similar to all-versus-all, whereas random cohorts struggles to minimize networks
without damaging their functionality. Random cohorts also maintains far fewer perfect networks than the other matchmakers. On this domain, we observe an approximate reduction in evaluation-estimation time from 2.4 seconds for baseline to 1.6 seconds for parents-versus-all; random cohorts ends up taking longer to evaluate due to increased network size.

The number of swaps in the best network and the errors for each matchmaking method seen in Figure~\ref{fig:sn} provide some insight behind the contrasting dynamics between these methods. Parents-versus-all produces significantly less error than random cohorts at the beginning of the run ($p\ll0.0001$ at 1e7 evaluations, Wilcoxon test; Glass's $\Delta=-2.42$), which results in a parents-versus-all finding the first perfect networks with much less swaps. The reduced starting size of perfect networks eases minimization and the lower error reduces the likelihood of selecting imperfect networks. Having more perfect networks in the population increases the chance of discovering smaller perfect networks, and so on. Finding smaller perfect networks becomes harder and harder, and eventually both the size of networks for parents-versus-all and all-versus-all plateau just above 70 swaps, significantly lower than random cohorts, which plateaus above 90 swaps ($p\ll0.00001$, Wilcoxon test; Glass's $\Delta=-7.67$).
\hypertarget{collision-game}{%
\subsection{Collision Game}\label{collision-game}}

\begin{figure*}
\centering
\begin{subfigure}[b]{\columnwidth}
\includegraphics[width=\columnwidth]{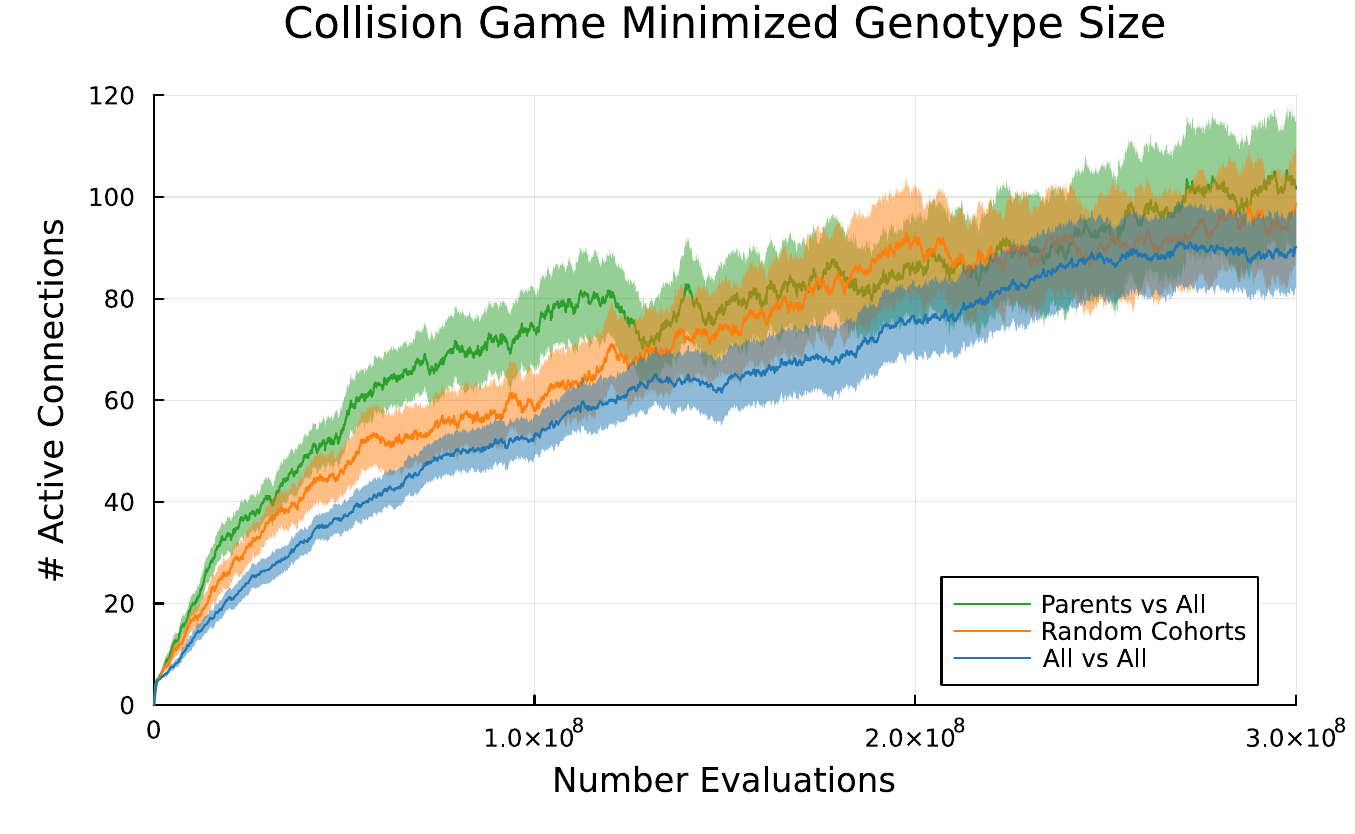}
\label{fig:cg-genotypesize}
\end{subfigure}
\begin{subfigure}[b]{\columnwidth}
\includegraphics[width=\columnwidth]{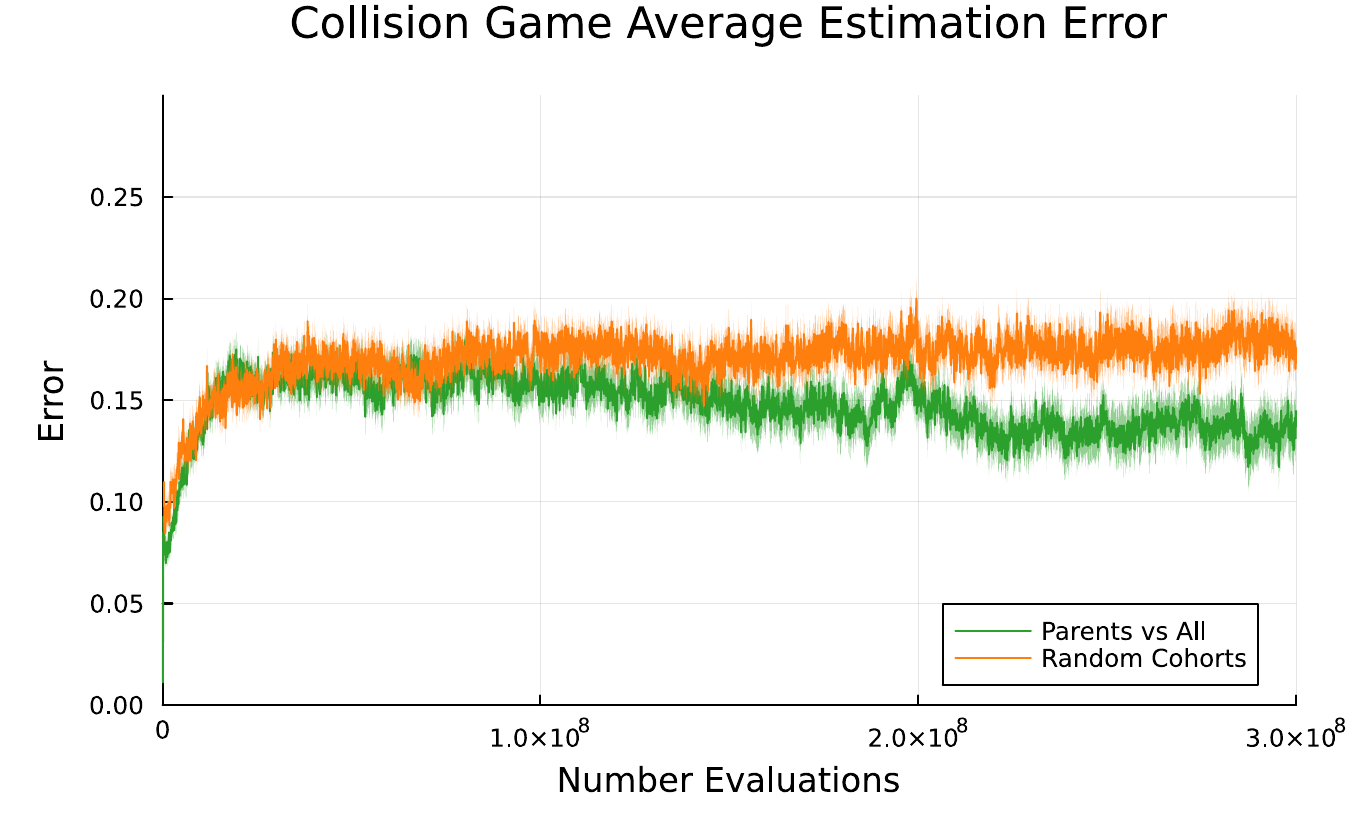}
\label{fig:cg-error}
\end{subfigure}
\caption{Collision Game results across thirty trials, 95\% confidence intervals shown. \textbf{Left}: Average number of connections across all minimized neural networks. Despite appearances, only parents-versus-all significantly accelerates growth during the first 50-100 million evaluations ($p<0.001$ at 5e7 evaluations, Wilcoxon test; Glass's $\Delta=1.3$), whereas random cohorts does not ($p<0.35$ at 5e7 evaluations, Wilcoxon test; Glass's $\Delta=0.48$) due to high variance between trials. All-versus-all eventually ``catches up'' in terms of the development by 3e8 evaluations (overlapping confidence intervals). Neither estimation methods performs worse than the baseline. \textbf{Right}: Estimator error. Minimum possible error is 0, maximum is 1. Despite appearances, parents-versus-all produces significantly lower error towards the beginning of the runs ($p<0.001$, Wilcoxon test; Glass's $\Delta=-1.02$) while errors near the end are not significantly different ($p<0.11$, Wilcoxon test; Glass's $\Delta=-0.57$).}\label{fig:cg-genotypesize}
\label{fig:cg}
\end{figure*}

Figure~\ref{fig:cg} shows complexity growth and estimation error for the parents-versus-all and random cohort regimes on the Collision Game. 
In prior work, control experiments have shown that networks are not biased to grow nor shrink, and that complexity generated as the result of adaptations to competitive and cooperative pressure. We hypothesized that our estimation techniques would accelerate the growth of these neural networks. While both methods in Figure \ref{fig:cg} demonstrate non-overlapping 95\% confidence intervals across 30 trials for the first 50 to 100 million evaluations, only parents-versus-all is statistically significant when compared to all-versus-all ($p<0.001$ at 5e7 evaluations, Wilcoxon test; Glass's $\Delta=1.3$), whereas random cohorts is not ($p<0.35$ at 5e7 evaluations, Wilcoxon test; Glass's $\Delta=0.48$) due to high variance across trials. As our networks become more complex, however, we observe that all-versus-all ``catches up'' to our estimation techniques in terms of complexity as the confidence intervals begin to overlap. 
We hypothesize that as complexity increases, adaptive mutations become rarer and thus require more evaluations to unearth. It may be that while our estimation techniques ``jump start'' the development of complexity, increasing levels of complexity become harder to obtain, allowing slower methods to eventually catch up under the same mutation scheme.

Configurations which develop complexity faster slow down sooner, making temporal comparison between configurations deceptive. At similar points in complexification, however, we observe a roughly 40\% decrease in time required for evaluation and estimation for per generation.

\hypertarget{discussion}{%
\section{Discussion}\label{discussion}}
Across all experimental settings, at least one of the proposed methods could approximate the dynamics of all-versus-all with substantially less computation, at least for the first hundred million evaluations or so.

Despite these successes, we observe the following shortcomings: (1) parents-versus-all matchmaking can introduce some bias, resulting in ``disconnected'' populations on CompareOnAll; (2) random cohorts struggles to minimize Sorting Networks when using a secondary fitness term; and (3) estimation initially accelerate the development of neural complexity in the Collision Game, but eventually estimation-free methods catch up.
These results are consistent with prior work in evolutionary settings that indicate estimation effectiveness varies by problem \citep{lalejini_phylogeny-informed_2023}. 
Additionally, we found that the optimal matchmaking scheme varies by problem as well. 

For systems that already incorporate phylogeny tracking for other purposes (e.g., \citep{dolson_quantifying_2018}), phylogeny-informed estimation adds little systematic complexity to achieve a significant reduction in the computation required to progress.
For domains where evaluation is expensive (e.g., evolutionary robotics) or that benefit from large populations (e.g., deep neuroevolution), our method can speed up existing all-versus-all algorithms without modifying the selection scheme or losing much information needed to preserve diversity.

The Collision Game results are nevertheless surprising.
Estimation performs as expected during the initial stages of the system, but we did not expect the naive all-versus-all approach to catch up when run for long enough on the open-ended domain. At worst, we expected a systemic collapse as seen in CompareOnAll.
We suspect that adaptive mutations for directly-encoded networks can be discovered in less evaluations for small networks, and may be rarer or require more evaluations to discover for large networks.
We hypothesize that mutation operators which efficiently discover adaptive mutations at high regions of complexity may allow our method to continue accelerating co-evolution on open-ended domains.

\hypertarget{future-work}{%
\section{Future Work}\label{future-work}}

The estimation methods proposed in this paper only work for algorithms that use asexual reproduction methods, as we leave phylogeny-informed fitness estimation in the context of sexual reproduction to future work. We also seek to apply these methods to deep neuroevolutionary domains, as these problems stand to benefit the most from interaction estimation. Additional promising directions lie in approaches that evolve modules \citep{angeline_evolutionary_1994, angeline_evolutionary_1993}
, leverage indirect encodings \citep{stanley_compositional_2007,stanley_hypercube-based_2009}, and generally scale better with complexity. We also believe related fields, such as multi-agent reinforcement learning, also stand to benefit from phylogeny-informed interaction estimation. \citep{majumdar_evolutionary_2020,long_evolutionary_2020,li_bridging_2024}. 

\hypertarget{conclusion}{%
\section{Conclusion}\label{conclusion}}

In this work, we demonstrated the viability of phylogeny-informed interaction estimation and matchmaking for accelerating co-evolutionary systems. Our findings reveal that these methods can approximate the dynamics of all-versus-all algorithms while significantly reducing the computation required, particularly in the early stages of search, but the optimal matchmaking strategy varies across domains. The Collision Game results suggest a diminishing return of estimation techniques as complexity increases over substantial periods of time, underscoring the necessity of testing both closed and open-ended domains to fully understand the implications and limitations of these methods.

\bibliographystyle{apalike}
\bibliography{main.bib}

\newpage

\appendix
\section{Appendix}

\begin{table}[h]
    \centering
    \begin{tabular}{c|c|c|c}
Domain & \# Parents & \# Children & Cohort Size \\
\hline
Numbers Games &  25 & 75 & 50 \\
Sorting Networks & 100 & 500 & 200 \\ 
Collision Game & 25 & 75 & 50
    \end{tabular}
    \caption{Hyperparameters for each domain}
    \label{tab:hparam}
\end{table}

\end{document}